\documentclass{article}
\usepackage{spconf,amsmath,graphicx}

\title{CS-NET: STRUCTURAL APPROACH TO TIME-SERIES FORECASTING FOR HIGH-DIMENSIONAL FEATURE SPACE DATA WITH LIMITED OBSERVATIONS}
\name{Weiyu Zong$^1$, Mingqian Feng$^2$, Griffin Heyrich$^1$, Peter Chin$^{1,3}$}
\address{$^1$Boston University Department of Mathematics and Computer Science\\
$^2$ Johns Hopkins University Department of Applied Mathematics and Statistics\\
$^3$ Dartmouth College Thayer School of Engineering}

\begin{document}
%

\maketitle
\begin{abstract}
In recent years, deep-learning-based approaches have been introduced to solving time-series forecasting-related problems. These novel methods have demonstrated impressive performance in univariate and low-dimensional multivariate time-series forecasting tasks. However, when these novel methods are used to handle high-dimensional multivariate forecasting problems, their performance is highly restricted by a practical training time and a reasonable GPU memory configuration. In this paper, inspired by a change of basis in the Hilbert space, we propose a flexible data feature extraction technique that excels in high-dimensional multivariate forecasting tasks. Our approach was originally developed for the National Science Foundation (NSF) Algorithms for Threat Detection (ATD) 2022 Challenge. Implemented using the attention mechanism and Convolutional Neural Networks (CNN) architecture, our method demonstrates great performance and compatibility. Our models trained on the GDELT Dataset finished 1st and 2nd places in the ATD sprint series and hold promise for other datasets for time series forecasting.

\end{abstract}

\section{Introduction}
\label{sec:intro}

With data of all types becoming more and more abundant, time series forecasting is taking on a more important role than ever in decision-making throughout many aspects of various domains, such as public health and safety\cite{1},  energy\cite{2}, transportation\cite{3}, and business\cite{4}. Recently, due to the rapid rise in complexity of time series data, a wide variety of machine learning models have been explored in forecasting time series. The methods include, but are not limited to, Temporal Convolutional Neural Network (TCNN), Gated Recurrent Unit (GRU), Long Short-Term Memory (LSTM), Elman Recurrent Neural Network (ERNN), and Multilayer Perceptron (MLP)\cite{5}. These models generally work well with low dimensional data with a high number of observations yet still suffer from numerous problems\cite{5}, and the performance of the models across different datasets has a large variance. Notably, current top performers like LSTM and GRU suffer from extremely high training and inference times compared to their counterparts. Existing CNN-based models like TCNN take the lead in their efficiency, but their performance drops drastically when handling data with high dimensionality. The trade-offs mentioned above lead us to rethink these approaches.

High-dimensional multivariate datasets undoubtedly pose challenges to existing forecasting models. However, they provide additional information that we can exploit. We make the assumption that there exist underlying interactions between different events occurring at different locations and times. Therefore, capturing the spatio-temporal relationships within the historical observations could help us interpret and forecast the states of future time frames, which are stochastic in nature. In this paper, we propose a novel strategy for feature extraction and neural network architecture design. Our strategy consists of four components--data pre-processing using structural decomposition, attention-based encoder,  convolution-based signal transform, and model ensemble. Each component is laid out in detail in the Methodology Section.

The goal of this research work is aligned with the ATD-2022 Challenge and its sponsors, NSF and National Geospatial-Intelligence Agency (NGA). We aim to develop algorithms that makes accurate predictions for high-dimensional time series data using limited historical observations which can be used in supporting public health and safety.

\section{Problem Formulation}
\label{sec:format}

The original problem that our model aimed to solve is proposed by the NSF ATD 2022 Challenge. Participants of the challenge are given a dataset derived from the GDELT project dataset. The GDELT project monitors print, broadcasts, and web media to record events across the globe and attribute them to state actors using the Conflict and Mediation Event Observations (CAMEO) coding system. In the CAMEO coding system, events are categorized into 20 distinct types according to their severity and rarity, ranging from ``Making Public Statements" to ``Engage In Unconventional Mass Violence." For example, on one side of the spectrum, events like ``Making Public Statements" could be happening at every moment around the world, which means that we have abundantly available past observations for these types of events. However, rare and catastrophic events, such as ``Engage In Unconventional Mass Violence," would have very limited available past observations. Organizers of ATD 2022 processed the raw GDELT data into a weekly-aggregated, geographic-region-level view.

The resulting dataset consists of the counts of the 20 CAMEO event types for 260 geographic regions across a 215-week window, which is a 215 by 5,200 table. The task is to use available past observations to forecast the next four states of the world in the future 4-week horizon. Namely, time series observations $T_{0, \dots, k}$ are given, where each $T_n \in N^{5,200}$ represents the state of the world at time $n$. Participants must produce a forecast function $$f: T_{0, \dots, k} \rightarrow T_{k+1, \dots, k+4}$$ where $T_{k+1, \dots, k+4}$ are the next four vectors representing future states of the world.

The backtesting procedure of ATD 2022 uses an expanding window starting with 100 weeks of past observations as the training set. The window expands for one observation at a time until the window exhausts the entire dataset. The consecutive 4 weeks of observations following the expanding window are then used as the testing set to evaluate the model performance.

\section{Methodology}

\subsection{Data Preprocessing: Structural Decomposition}

In our context, it is safe to assume that a sequence of numbers, such as a time series dataset, often represents interpretable measurements from the real world. Given this assumption, we propose that for any high-dimensional longitudinal dataset, we consider the key structural dimensions and arrange the data set accordingly, which can help deep-learning models converge faster. For example, we can decompose the ATD version of the GDELT Dataset into three major dimensions–Time, Event types, and Region. As illustrated in figure 1, we obtained a cuboid-shaped data block using the proposed approach, where each dimension corresponds to time, event categories, and space, respectively. In other words, every cross-section of the cuboid in the spatial dimension is a separate panel data of time and event types.
 
\subsection{Attention-based encoder}

In many situations, we do not necessarily have prior knowledge about the relationships between different time series in our dataset. Inspired by the success of attention-based structures in natural language processing, we included the attention-based encoder in our model. We aim to use this encoding layer as an additional effort to capture potential connections across the different time series. In our case, as shown in Figure 2, we applied an attention-based encoder throughout the entire spatial dimension to emphasize underlying associations across different geographic locations. The encoder layer is the same as the structure proposed in the original paper \cite{6}.

\begin{figure}[htb]

\begin{minipage}[b]{1.0\linewidth}
  \centerline{\includegraphics[width=8.5cm]{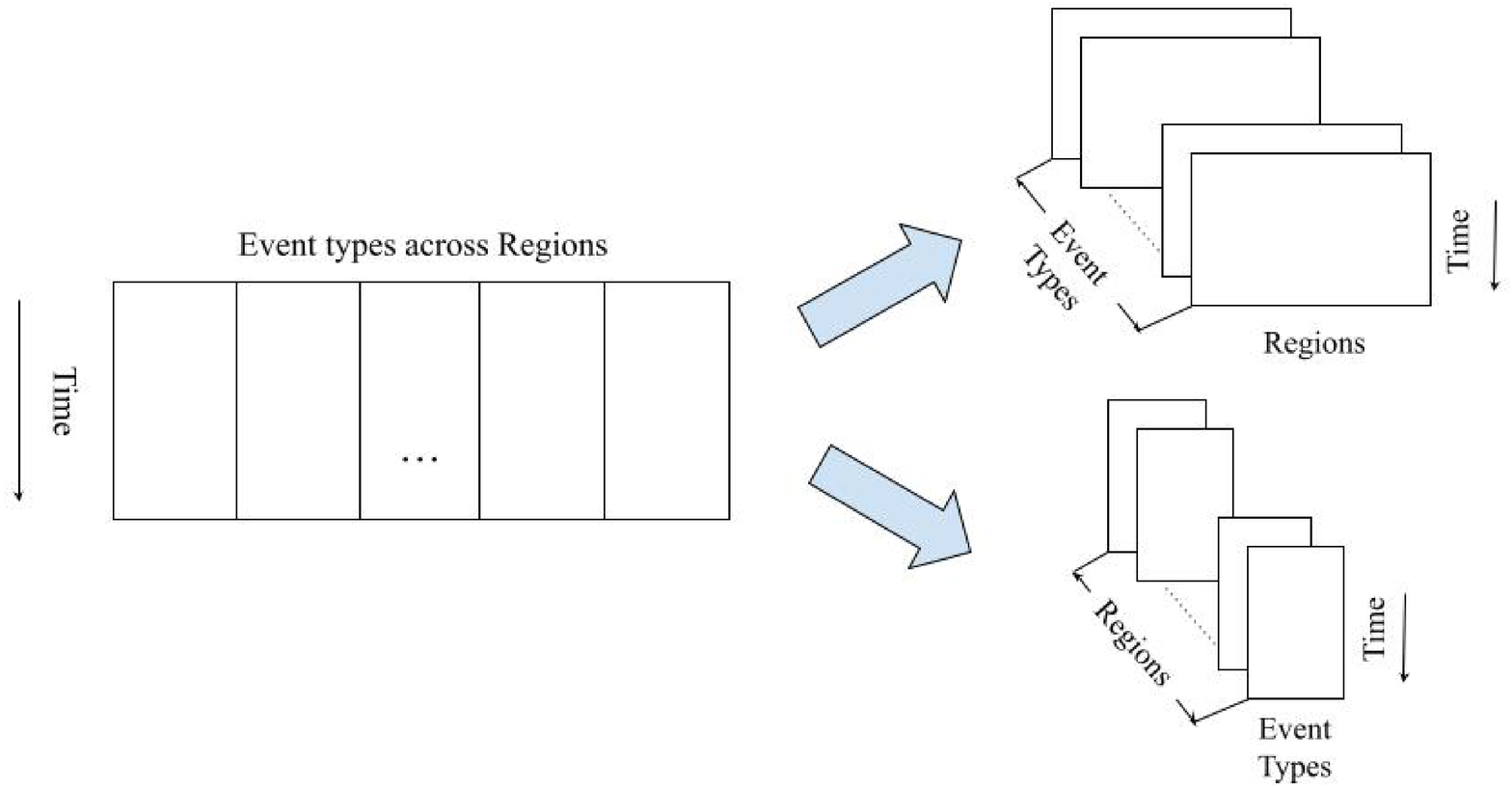}}
  {\textbf{Figure 1}: The proposed method for data preprocessing--Structural Decomposition. The input dataset is decomposed into two cuboid-shaped data blocks consisting of layers of panel data. In the ATD Dataset the panels are  ``Time" vs. ``Regions" and ``Time" vs. ``Event types." Note that different datasets can be decomposed differently according to the Key Dimensions.}
\end{minipage}

\end{figure}

\subsection{Convolution-based signal transform}

Signal transform techniques like the Fourier transform are very commonly used in signal processing. It is capable of revealing essential characteristics of a signal by approximating the signal as a linear combination of its basis frequencies. Previous research suggests that the use of a windowed Fourier transform enables a better interpretation of the randomness in a given signal \cite{7}. Notice that when the input signal is projected into an arbitrary Hilbert space, the Fourier transform operation can be interpreted as a change of basis. For example, suppose that the input signal F is in time domain t; we can write the following to express a windowed Fourier transform, $$F = \langle \vec{f}, e^{i\omega t} \rangle =  \int_{-\infty}^{\infty} f(t) g(t-s) e^{i\omega t} \, dt$$ where $f$ is the input signal, $g$ is a window function, and $e^{i\omega t}$ is the family of trigonometric functions serving as the orthonormal basis in the Hilbert space.

To expand on this idea, we need to produce a transformation that can decompose our panel of time series into a set of basis, which we can use as features to make combinations and generate predictions. In our case, every cross-section of the reshaped cuboid data block is a 2-D panel. For instance, if we take a slice of the spatial dimension, we will obtain a time versus event type panel. Inspired by the work in \cite{16}, image representation using 2-D Gabor Wavelets, we realized that applying the concept mentioned above on this slice of data can be implemented efficiently using a 2-D convolution. i.e.,  $$F_{n}(t,s) =  \sum_{x}\sum_{y} f(x,y) k_n (t-x,s-y)\ $$ where $k_n \in K$ is $n^{\text{th}}$ kernel of the set of kernels (basis) to be learned in the training process. As shown in figure 2, these extracted features are fused through an MLP layer to generate predictions.

\begin{figure}[htb]

\begin{minipage}[b]{1.0\linewidth}
  \centerline{\includegraphics[width=8.5cm]{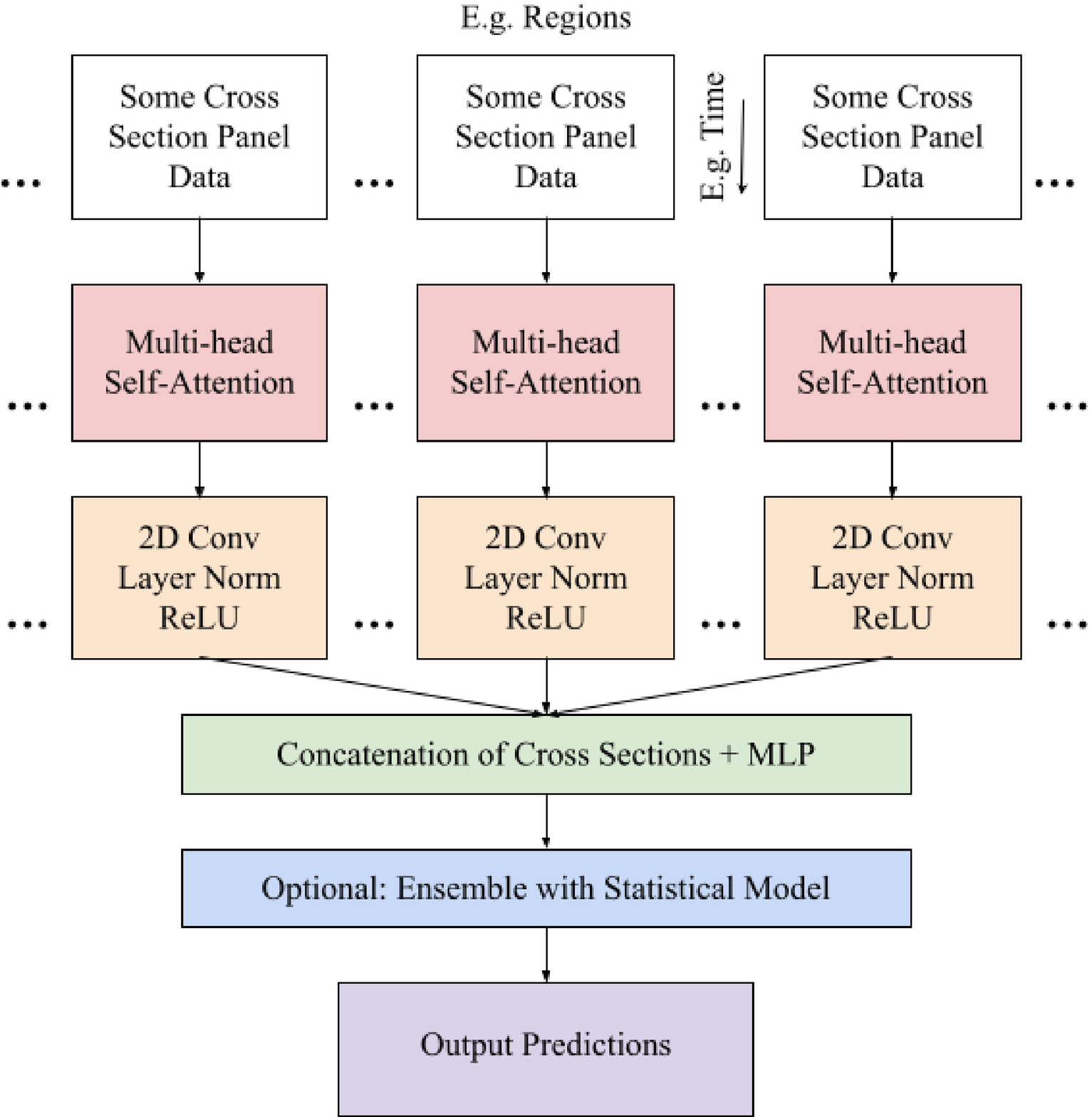}}
  {\textbf{Figure 2}: An overview of the proposed architectural design }
 
\end{minipage}
\end{figure}

\subsection{Optional Model Ensemble Layer}

During the model development process, we noticed that statistical models like ARIMA models tend to have a higher bias, making them better at following the general trend in the time series data. However, machine learning-based models are more likely to have a higher variance, especially when given limited observations as in our case. As a result, machine learning-based methods are better at predicting localized and drastic changes in the data. Depending on the specific dataset in consideration, we propose an optional layer that performs a weighted sum of our model predictions and additional statistical model predictions to balance the bias-variance tradeoff problems. In our case, we ensemble our model predictions with predictions from an additional Vector Auto-regression model (VAR).

\subsection{Metrics}

Since the ATD version of the GDELT Dataset contains a total of 5,200 time series with distinct scales and the values of some observations are zero, we need a metric invariant to the magnitude of data points and is capable of handling zero as truth values. Therefore, we chose Mean Absolute Scaled Error (MASE) as our primary metric. MASE is defined as the ratio between the mean absolute error of the forecast values and the mean absolute error of the in-sample one-step naive forecast \cite{8}. i.e.,
$$MASE = \sum_{k=1}^{h} \frac {1} {h} \frac {|y_{t+k}-\hat{y}_{t+k}|}{(n-1)^{-1} \sum_{i=2}^{n} |y_i-y_{i-1}|} $$ where $h$ is the length of the forecasting horizon; $n$ is the total number of observations; $\hat{y}$ is the predicted value for observation $y$.

Besides the accuracy of the model, we also want to compare different model's capabilities in avoiding making large forecasting errors. Thus, we have also included Mean Squared Error (MSE) as a secondary reference since MSE, as defined in the expression below, emphasizes the large forecasting errors. $$MSE = \frac{1} {n} \sum_{i=1}^{n} (y_i-\hat{y_i})^2$$ where $n$ is the total number of observations; $\hat{y}$ is the predicted value for truth value $y$.

\section{Experimental Setup}

\subsection{Baselines for model comparisons}

To assess the performance of our proposed method, we are using three state-of-the-art deep learning models and one classic statistical model for time series forecasting as baselines. Namely, Google AI's Temporal Fusion Transformer (TFT) \cite{10}, N-Beats Forecaster from Element AI \cite{11}, DeepAR Forecaster from Amazon Research \cite{12}, and Vector Autoregression (VAR) Model. The three deep learning models were implemented with GluonTS, a Python library \cite{13}, and the VAR model was implemented using Python Statsmodels library \cite{14}. The parameters for each model are chosen carefully using grid search method around the default values recommended by the library implementations to achieve their best performance.

\subsection{Dataset preparation and training details}

The performance of our proposed model is demonstrated on two separate datasets. 1) ATD 2022 Dataset, 2) Wikipedia Web Traffic Dataset. Dataset 1 was given to participants for model development. The details of our proposed Structural Decomposition procedure on Dataset 1 are already described in the previous sections, which we do not repeat here. Dataset 2 is obtained from a Kaggle competition \cite{15} with the goal of predicting future web traffic of Wikipedia web pages. The raw dataset contains 145, 063 time series. We randomly selected a subset of 1,400 distinct time series of web pages for active keywords in multiple languages. Each of these is the recorded daily web traffic spanning 500 days, resulting in a 500 by 1,400 table as the input. In addition, the name of each time series also contains information regarding the language a web page is written in, the device type users used to access a web page, and keywords related to a web page.  The key structural dimensions that we are considering here are Time, User Device types, Language, and Keyword categories.

For Dataset 1, all models are trained on a sliding window of 170 observations in size to predict the next 8 observations. Similarly, for Dataset 2, we trained the models on a sliding input window of size equal to 300 observations to predict the next 4 observations. Note that we choose different training window sizes because Dataset 2 has more number of observations. In addition, our model has already demonstrated strong performance in the ATD 2022 challenge with a forecast horizon equal to 4. So we decided to test our model's performance over a longer forecast horizon.

\begin{table}

    \centering
    \begin{tabular}{|c||c|c|c|c|}
        \hline
         &  \multicolumn{2}{|c|}{ATD Data}  &  \multicolumn{2}{|c|}{Wikipedia Data}\\
        \hline
        Model Names & MASE & MSE & MASE & MSE\\
        \hline
        CS-Net 1 & 1.126 & 73.90 & 0.773 & 785.47 \\
        CS-Net 2 & 1.032 & 61.08 & 0.731 & 736.19 \\
        \textbf{CS-Net 3} & \textbf{1.015} & \textbf{60.50} & \textbf{0.694} & \textbf{718.19} \\
        DeepAR & 1.136 & 74.04 & 0.955 & 964.64\\
        TFT & 1.078 & 63.43 & 0.752 & 748.01 \\
        VAR & 1.193 & 68.19 & 0.977& 973.69 \\
        N-Beats & 1.112 & 73.43 & 0.748& 746.44 \\
        \hline
    \end{tabular}
    
    \caption{Performance of different models on two datasets using MASE and MSE as metrics}
    \label{tab:my_label}
    

\end{table}

\section{Results and Analysis}

\subsection{Ablation and Comparison Analysis}

We made three variations of our model to quantify the effectiveness of each of our proposed architectural components. For simplicity, we named our proposed model design Cross-Sectional-Net (CS-Net). CS-Net 1, 2, and 3 correspond to the three variations. CS-Net 1 only has the Convolution-based signal transform. Predictions from CS-Net 2 are a weighted sum of CS-Net 1 outputs and predictions made by a VAR forecaster. Lastly, CS-Net 3 has all the components proposed in the Methodology section, i.e., Convolution based signal transform, Attention-based encoder, and Model ensemble.

According to Table \ref{tab:my_label}, we see that regardless of the dataset or metric used, the proposed Convolution-based signal transform across data sections alone yield a relatively competitive performance. Its metric scores is close to that of TFT and N-Beats forecasters.
The inclusion of a statistical model ensemble further improved the proposed method's performance, which verified our hypothesis that variances and biases need to be balanced. It is worth noting that even though classical statistics models like VAR do not show great performance when used alone, they can be helpful in supporting the robustness of deep learning-based predictions.

We also noticed that in the ATD Dataset, the improvements from adding multi-head self-attention layer is not as significant as in Wikipedia Traffic Dataset. ATD Dataset has less data for training, so models may have over-fitted to some time series and not have fully converged in others. Thus, we used regularization techniques, such as dropout and adding regularization in the Adam loss function. We also expect more notable improvements given relatively more historical observations. 

With all of the proposed components integrated, CS-Net 3 outperforms the other baseline models in both datasets. Particularly, it takes a strong lead in the MSE metric, indicating that it successfully avoided making large prediction errors.

\subsection{Limitations}

Despite that the proposed method outperforms state-out-the-art architectures like TFT, N-Beats and DeepAR Forecaster, the proposed method relies on a stronger assumption on a given dataset that it must contain some type of structure. This assumption reduces its generalizability on different datasets. In addition, TFT provides interpretability of variable importance, which could be useful in applications.

\section{Conclusions}

In this paper, we adopted a novel approach to handle high-dimensional multivariate time series forecasting tasks with limited availability of historical observations. The proposed approach has shown success in the NSF ATD 2022 Challenge. This method also outperforms other cutting-edge methods in time series forecasting in subsequent experiments on additional datasets,   confirming the effectiveness of the proposed method. This research work can potentially be used in supporting fields of public health and safety.

\section{Acknowledgement}

This work was funded by the United States National Science Foundation Division of Mathematical Sciences, under the award grant NSF-DMS 1737897. The authors would like to thank Penn State Applied Research Laboratory for organizing the NSF ATD 2022 Challenge. All authors contributed equally to this work. Listing order is random.

\vfill\pagebreak

\bibliographystyle{IEEEbib}

\bibliography{references}

\begin{thebibliography}{10}

\bibitem{1}
C.~Vladescu V.~Olsavszky, M.~Dosius and J.~Benecke,
\newblock ``Time series analysis and forecasting with automated machine
  learning on a national icd-10 database,''
\newblock {\em International journal of environmental research and public
  health}, vol. 17, pp. 1, July 2020.

\bibitem{2}
Aditya Ashok, Manimaran Govindarasu, and Venkataramana Ajjarapu,
\newblock ``Online detection of stealthy false data injection attacks in power
  system state estimation,''
\newblock {\em IEEE Transactions on Smart Grid}, vol. 9, no. 3, pp. 1636--1646,
  2018.

\bibitem{3}
S.~L. Dhingra, P.~P. Mujumdar, and Rajesh~H. Gajjar,
\newblock ``Application of time series techniques for forecasting truck traffic
  attracted by the bombay metropolitan region,''
\newblock {\em Journal of Advanced Transportation}, vol. 27, no. 3, pp.
  227--249, 1993.

\bibitem{4}
A.~Dorestani Z.~Rezaee and S.~Aliabadi,
\newblock ``Application of time series analyses in big data: Practical,
  research, and education implications,''
\newblock {\em Allen Press}, vol. 15, November 2017.

\bibitem{5}
Pedro Lara-Ben\'{\i}tez, Manuel Carranza-Garc\'{\i}a, and Jos\'{e}~C. Riquelme,
\newblock ``An experimental review on deep learning architectures for time
  series forecasting,''
\newblock {\em International Journal of Neural Systems}, vol. 31, no. 03, pp.
  2130001, 2021,
\newblock PMID: 33588711.

\bibitem{6}
Ashish Vaswani, Noam Shazeer, Niki Parmar, Jakob Uszkoreit, Llion Jones,
  Aidan~N Gomez, \L~ukasz Kaiser, and Illia Polosukhin,
\newblock ``Attention is all you need,''
\newblock in {\em Advances in Neural Information Processing Systems}, I.~Guyon,
  U.~Von Luxburg, S.~Bengio, H.~Wallach, R.~Fergus, S.~Vishwanathan, and
  R.~Garnett, Eds. 2017, vol.~30, Curran Associates, Inc.

\bibitem{7}
Véronique Millette and Natalie Baddour,
\newblock ``Signal processing of heart signals for the quantification of
  non-deterministic events - biomedical engineering online,'' Jan 2011.

\bibitem{16}
Tai~Sing Lee,
\newblock ``Image representation using 2d gabor wavelets,''
\newblock {\em IEEE Transactions on Pattern Analysis and Machine Intelligence},
  vol. 18, no. 10, pp. 959--971, 1996.

\bibitem{8}
Rob~J. Hyndman and Anne~B. Koehler,
\newblock ``Another look at measures of forecast accuracy,''
\newblock {\em International Journal of Forecasting}, vol. 22, no. 4, pp.
  679--688, 2006.

\bibitem{10}
Bryan Lim, Sercan~Ö. Arık, Nicolas Loeff, and Tomas Pfister,
\newblock ``Temporal fusion transformers for interpretable multi-horizon time
  series forecasting,''
\newblock {\em International Journal of Forecasting}, vol. 37, no. 4, pp.
  1748--1764, 2021.

\bibitem{11}
Boris~N. Oreshkin, Dmitri Carpov, Nicolas Chapados, and Yoshua Bengio,
\newblock ``N-beats: Neural basis expansion analysis for interpretable time
  series forecasting,'' 2019.

\bibitem{12}
David Salinas, Valentin Flunkert, Jan Gasthaus, and Tim Januschowski,
\newblock ``Deepar: Probabilistic forecasting with autoregressive recurrent
  networks,''
\newblock {\em International Journal of Forecasting}, vol. 36, no. 3, pp.
  1181--1191, 2020.

\bibitem{13}
Alexander Alexandrov, Konstantinos Benidis, Michael Bohlke-Schneider, Valentin
  Flunkert, Jan Gasthaus, Tim Januschowski, Danielle~C. Maddix, Syama
  Rangapuram, David Salinas, Jasper Schulz, Lorenzo Stella, Ali~Caner Türkmen,
  and Yuyang Wang,
\newblock ``{GluonTS: Probabilistic and Neural Time Series Modeling in
  Python},''
\newblock {\em Journal of Machine Learning Research}, vol. 21, no. 116, pp.
  1--6, 2020.

\bibitem{14}
Skipper Seabold and Josef Perktold,
\newblock ``statsmodels: Econometric and statistical modeling with python,''
\newblock in {\em 9th Python in Science Conference}, 2010.

\bibitem{15}
Google,
\newblock ``Web traffic time series forecasting,'' 2017.

\end{thebibliography}

\end{document}